\documentclass[11pt,a4paper]{article}
\usepackage[T1]{fontenc}
\usepackage{mathptmx}
\usepackage{amsmath,amssymb,booktabs,graphicx,float,placeins}
\usepackage[margin=25mm]{geometry}
\usepackage[numbers,sort&compress]{natbib}
\usepackage[hidelinks]{hyperref}
\usepackage{caption}
\newcommand{\Real}{\mathbb{R}}

\title{JAX-WAVE: A Differentiable Physics-Supervised Framework for Inverse Wave Design in Heterogeneous Media}
\author{Honghui Du$^{1,*}$, Binyao Guo$^1$, Yushu Pan$^2$}
\date{}
\begin{document}
\maketitle
\noindent $^1$ Department of Civil, Environmental, and Geo- Engineering, University of Minnesota, Minneapolis, MN, USA.\\
$^2$ Department of Computer Science, Columbia University, New York, NY, USA.\\
$^*$ Corresponding author: \href{mailto:du000211@umn.edu}{du000211@umn.edu}.\\
Author emails: du000211@umn.edu; guo00544@umn.edu; yp2602@columbia.edu.
\begin{abstract}
Designing a prescribed wave field in a heterogeneous medium requires an initial pressure distribution that compensates for the medium's propagation response. Repeated numerical inversion solves this task afresh for each target and environment. We introduce JAX-WAVE for amortized inverse wave design: a shared neural network maps desired terminal fields and known wave-speed maps directly to bounded initial pressures. Training through a differentiable finite-difference time-domain solver uses the physical design objective without requiring optimized control labels. Linear propagation and symmetric pressure bounds imply an exact sign-reversal relation for the regularized optimum. A shared-weight projection enforces this relation, preserves feasibility, and satisfies an exact identity for the paired-sign objective. We characterize the discrete propagator and its weighted adjoint to connect learning, numerical inversion, and optimality assessment. On an independent confirmation set, three trained models achieve unseen-material errors of 25.24\% for geometric targets and 4.77\% for reachable targets. Matched-capacity comparisons show improved reconstruction with the sign parameterization. Within a 5\,ms median online budget, the learned map reduces mean error by 45\% relative to accelerated numerical inversion on the heterogeneous benchmark; longer numerical solves recover greater accuracy. The results establish a reusable inverse-design procedure and show how its accuracy and computational value depend on target structure and query workload.

\end{abstract}
\section{Introduction}\label{sec:intro}
A desired wave pattern does not directly specify the excitation needed to create it. The initial pressure evolves through the medium, and its positive and negative contributions combine at the design time. When the wave speed varies in space, this evolution changes with the material configuration. Inverse wave design must therefore determine an excitation jointly from the target field and the propagation environment. Time-reversal focusing \cite{fink1992reversal} and acoustic holograms \cite{melde2016holograms} illustrate how propagation can be used to shape a pressure field. Learned hologram generation and heterogeneous physics constraints further connect desired patterns to source designs \cite{lee2022,e2ehpc2026}.

The physical difficulty lies in coordinating this evolution. A localized terminal feature may require a spatially extended initial pattern, with signed contributions that reinforce one another at the intended location and cancel elsewhere. An amplitude limit restricts how strongly the control can compensate for weak propagation responses. Consequently, a target drawn as a geometric image need not be as easy to reproduce as a field generated by an admissible excitation. These features make initial-pressure design a useful setting for studying how a learned inverse model handles propagation, heterogeneous materials, and control constraints together.

For a known medium, numerical optimization recovers an initial pressure by alternating forward propagation with gradient-based control updates. Adjoint methods make these gradients accessible without assembling a full sensitivity matrix \cite{plessix2006adjoint}. Differentiable acoustic optimization \cite{diffpat2021} expresses the design process through a computational model. j-Wave provides differentiable acoustic solvers in JAX and demonstrates their use in inverse modeling and constrained source optimization \cite{stanziola2023jwave}. Related tools support finite-element analysis \cite{xue2023jaxfem}, crystal-plasticity inverse design \cite{hu2025jaxcpfem}, and material-point simulation \cite{du2026jaxmpm}. Such solvers allow the designer to refine a control until a prescribed accuracy or computation limit is reached. For repeated design, we seek to reuse this computation as targets and material realizations change.

We address this task with a conditional inverse map that takes the desired field and medium as inputs and predicts the initial control directly. Learning this map distributes computation across future design queries. The design objective concerns the propagated field, so the forward wave model must guide the predicted control. The target distribution also matters: the model must approximate a family of constrained inverse solutions whose difficulty changes with both spatial structure and propagation response.

Physics-supervised learning provides a way to construct this map without first optimizing every training example to obtain a control label. Physical information can enter a model through its data, objective, or architecture \cite{pateras2023taxonomy}. Physics-informed neural networks use governing-equation residuals to approximate solutions \cite{raissi2019pinn}, including density-driven flow in porous media \cite{du2023co2pinn}. For initial-pressure design, a time-domain propagator gives a particularly direct training signal: it evolves the proposed excitation, measures the terminal mismatch, and returns the sensitivity of that mismatch to the network.

Differentiable simulation has already supplied training signals for control \cite{hu2020difftaichi,holl2020control} and learned solver corrections \cite{um2020solver}. Shared function-to-function maps are central to DeepONet and Fourier neural operators \cite{lu2021deeponet,li2021fno}, with history-aware operators extending the idea to path-dependent material response \cite{guo2025hano}. A learned forward operator can also be embedded in a design optimization loop \cite{wang2021optimization}. In the present setting, the shared map predicts the initial control itself, and the numerical wave solver supplies its training objective. This choice retains the propagation model while making control generation a network evaluation after training.

Related acoustic studies address several parts of this task. Physics-based training has been used to learn hologram reconstruction without paired design labels \cite{physnet2022}; transcranial models predict pressure fields and phase corrections \cite{tusnet2026}; and a learned initial-pressure representation can be fitted to observations through a differentiable solver \cite{verburg2025}. Our focus is the combined requirement of repeated prediction across targets and known heterogeneous media, pointwise feasibility of the output, and fidelity to a prescribed terminal field. A computational framework for this setting should connect the representation being learned to the constrained inverse problem that defines a successful design.

That inverse problem supplies useful structure. Linear propagation, a quadratic control penalty, and symmetric amplitude bounds imply that reversing the target reverses the unique optimal control. Physically, changing compression to rarefaction reverses the wave field without changing the material. The same relation therefore holds across all speed maps in the model. Group averaging provides an established construction for imposing such symmetries \cite{cohen2016group,sannai2024reynolds}. Here, its two-element form preserves the pressure bounds and admits an exact quadratic identity for the paired target objective. This gives a mathematical characterization of the projection and a specific physical relation to test in matched training comparisons.

\begin{figure}[H]
\centering\includegraphics[width=\textwidth]{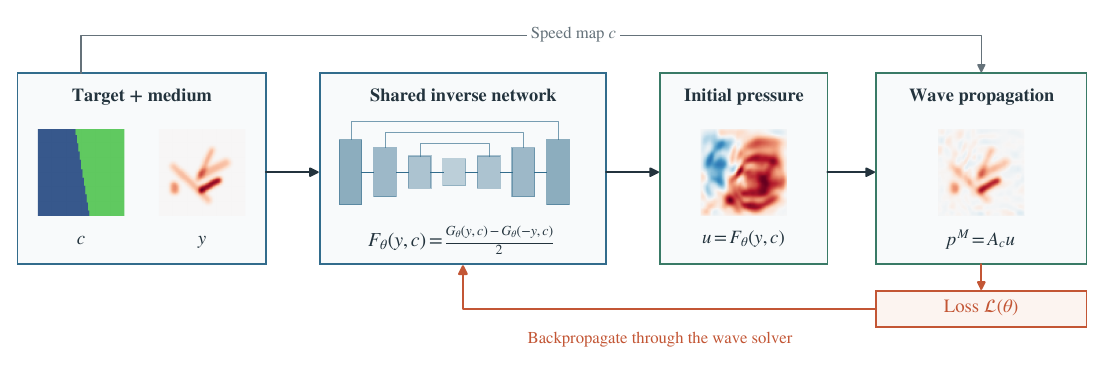}
\caption{JAX-WAVE training and prediction. The desired terminal field and known speed map condition a bounded initial-pressure prediction. During training, the FDTD solver propagates this control and supplies the field loss; gradients pass through the solver to the network. During prediction, the trained network generates the control directly. A subsequent propagation can evaluate the resulting field.}
\label{fig:workflow}
\end{figure}

We introduce JAX-WAVE as a physics-supervised computational framework for repeated inverse wave design (Figure~\ref{fig:workflow}). Its contributions are:
\begin{enumerate}
\item A shared inverse-design model trained through differentiable finite-difference time-domain (FDTD) propagation. The model conditions on the target and known heterogeneous medium and predicts bounded initial pressures without optimized controls as training labels.
\item A feasible sign-symmetric parameterization with an exact paired-objective identity. We derive the sign relation of the constrained optimum, impose it through shared-weight group averaging, and evaluate its training benefit with matched capacity, initialization, and minibatch sequences.
\item A computational assessment linking material generalization and propagation refinement to numerical accuracy references and online cost. We quantify reconstruction accuracy across targets and media, the effect of sign symmetry, and the trade-off between reconstruction error and prediction time.
\end{enumerate}
The formulation connects the discrete propagation operator, its weighted adjoint, and the constrained objective before presenting the learning procedure. Numerical experiments then examine the accuracy of the resulting controls and the computational value of reusing the trained map.

\section{Computational Formulation and Inverse Model}\label{sec:method}
\subsection{Wave Model and Design Variables}
Let $p(\mathbf{x},t)$ denote dimensionless pressure in the unit square $\Omega=(0,1)^2$, with prescribed positive wave speed $c(\mathbf{x})$. We consider
\begin{align}
 p_{tt}&=c(\mathbf{x})^2\Delta p &&\quad\text{ in }\Omega,\quad 0<t\leq T,\label{eq:wave}\\
 p&=0 &&\quad\text{ on }\partial\Omega,\nonumber\\
 p(\mathbf{x},0)&=u(\mathbf{x}),\qquad p_t(\mathbf{x},0)=0.\label{eq:initial}
\end{align}
The design variable is the initial pressure $u$; the target $y$ specifies the pressure at time $T$. With particle velocity $v$, constant density $\rho$, and spatially varying bulk modulus $K$, the first-order acoustic equations $p_t=-K\nabla\cdot v$ and $\rho v_t=-\nabla p$ give Equation~\eqref{eq:wave} with $c^2=K/\rho$. Thus, the speed multiplies the Laplacian in this model. The homogeneous boundary values and zero initial derivative are fixed across all design queries.

On an $n\times n$ interior grid, admissible controls belong to
\begin{equation}
 \mathcal C=\{u\in\Real^{n^2}: |u_i|\leq1\ \text{for all }i\}.\label{eq:box}
\end{equation}
\subsection{Discrete Propagation, Stability, and Adjoint}
Let $D_h$ be the symmetric five-point Dirichlet Laplacian, $h=1/(n+1)$, $C=\operatorname{diag}(c)$, and $\Delta t=T/M$. Centered finite differences \cite{langtangen2017fd} give the discrete propagation
\begin{align}
 p^0&=u,\qquad p^1=u+\tfrac12\Delta t^2 C^2D_hu,\label{eq:start}\\
 p^{k+1}&=2p^k-p^{k-1}+\Delta t^2C^2D_hp^k.\label{eq:fdtd}
\end{align}
The Taylor start imposes the initial derivative condition. We denote the terminal operator by $A_cu=p^M$. Although heterogeneous speeds couple spatial modes, $A_c$ remains linear in the initial pressure for each fixed medium.

We impose the sufficient two-dimensional Courant condition $\Delta t\,c_{\max}/h\leq1/\sqrt{2}$ for stable propagation.

The terminal map has a useful algebraic structure. Set $S=CD_hC$, which is symmetric, and let $T_M$ denote the degree-$M$ Chebyshev polynomial of the first kind. The recurrence gives
\begin{equation}
 A_c=CP C^{-1},\qquad P=T_M\!\left(I+\frac{\Delta t^2}{2}S\right)=P^\top.\label{eq:discrete_operator}
\end{equation}
Under the stated Courant condition, the argument of $T_M$ has spectrum in $[-1,1]$. Hence
\begin{equation}
 \|P\|_2\leq1,\qquad \|A_c\|_2\leq\frac{c_{\max}}{c_{\min}}.\label{eq:operator_bound}
\end{equation}
The similarity transform also yields the weighted adjoint
\begin{equation}
 A_c^\top r=C^{-2}A_c(C^2r).\label{eq:adjoint}
\end{equation}
Thus, the same propagation routine supplies the forward response and its adjoint after diagonal speed weighting. This structure supports both gradient-based numerical inversion and an independent check of the differentiated solver. Supplementary Material S1 reports a numerical comparison with automatic differentiation. Appendix~\ref{app:operator} gives the polynomial derivation and optimality certificate.

\subsection{A Feasible Inverse Map with Exact Sign Symmetry}\label{sec:symmetry}
For each target--medium pair, the control problem uses a quadratic Tikhonov penalty \cite{hansen2010inverse}:
\begin{equation}
 \min_{u\in\mathcal C} J_c(u;y),\qquad
 J_c(u;y)=\|A_cu-y\|_2^2+\lambda\|u\|_2^2,\quad\lambda>0.\label{eq:inverse}
\end{equation}
The Hessian $2(A_c^\top A_c+\lambda I)$ is positive definite, so the minimizer $u_\star(y,c)$ is unique. Since the objective and feasible set satisfy $J_c(-u;-y)=J_c(u;y)$ and $\mathcal C=-\mathcal C$, the inverse map obeys
\begin{equation}
 u_\star(-y,c)=-u_\star(y,c).\label{eq:optimal_sign}
\end{equation}

We represent the base map $G_\theta$ with a U-Net \cite{ronneberger2015unet} that receives the target and medium through two input channels, $[y,c-1]$. Its encoder--decoder architecture combines spatial scales through skip connections, and a final $\tanh$ activation bounds the predicted pressure. Section~\ref{sec:implementation} specifies the network configurations used in the experiments.

We impose Equation~\eqref{eq:optimal_sign} by the shared-weight projection
\begin{equation}
 F_\theta(y,c)=\tfrac12\bigl[G_\theta(y,c)-G_\theta(-y,c)\bigr].\label{eq:sym}
\end{equation}
This is a two-element group averaging construction \cite{cohen2016group,sannai2024reynolds}. It preserves the box constraint, is odd in $y$, and gives $F_\theta(0,c)=0$. It adds evaluations of the base network but no trainable parameters. The two signs can also be processed as a batch.

The projection has an exact objective identity. Set $a=G_\theta(y,c)$ and $b=G_\theta(-y,c)$. Expanding the quadratic objective gives
\begin{align}
 J_c\!\left(\frac{a-b}{2};y\right)
 ={}&\frac{J_c(a;y)+J_c(b;-y)}{2}\nonumber\\
 &-\frac{\|A_c(a+b)\|_2^2+\lambda\|a+b\|_2^2}{4}.\label{eq:sym_identity}
\end{align}
The final term is nonnegative. Projection therefore cannot increase the average objective over a target and its negative for a fixed base network. Reconstruction gains on nonnegative geometric targets are evaluated through the matched training comparisons in Section~\ref{sec:symmetry_results}.

\subsection{Training through Propagation}
Training couples the predicted control to a numerical model of its physical response. This integration of neural representations with discretized physics also underlies neural-integrated meshfree methods \cite{du2024nim}. Here, the propagator evaluates how closely the control reproduces the target. For a minibatch $\mathcal B$ of nonzero targets and speed maps, the loss is
\begin{equation}
 \mathcal L(\theta)=\frac{1}{|\mathcal B|}\sum_{(y,c)\in\mathcal B}
 \frac{\|A_cF_\theta(y,c)-y\|_2^2+\lambda\|F_\theta(y,c)\|_2^2}{\|y\|_2^2}.\label{eq:loss}
\end{equation}
The normalization gives each example a relative field-error scale. Uniform-grid quadrature weights cancel in this ratio. We minimize this loss with Adam \cite{kingma2015adam}, differentiating through the network and all propagation steps. No optimized controls are supplied as labels. Physics-informed DeepONets likewise use governing equations to train shared maps without paired solution observations \cite{wang2021physicsdeeponet}; Equation~\eqref{eq:loss} applies physical supervision to control prediction through the discrete propagator.

At deployment, evaluating $F_\theta$ on a target and its known speed map returns an admissible initial pressure. A forward solve can then verify the corresponding terminal field.

\subsection{Numerical Inversion and Accuracy References}\label{sec:numerical}
Projected gradient, the fast iterative shrinkage-thresholding algorithm (FISTA) \cite{beck2009fista}, and gradient-restarted FISTA \cite{odonoghue2015restart} solve Equation~\eqref{eq:inverse} with the same box bounds. Adjoint methods evaluate gradients without assembling the full forward Jacobian \cite{plessix2006adjoint}. For half of the objective, the gradient is
\begin{equation}
 g(u)=A_c^\top(A_cu-y)+\lambda u.\label{eq:gradient}
\end{equation}
Using Equation~\eqref{eq:adjoint}, evaluating a gradient requires one propagation of the control and one weighted propagation of the residual. Each update clips the candidate to $[-1,1]$. Backtracking begins at $L=1$ and doubles it until quadratic majorization holds, then retains the accepted bound for the next iteration. Timings include all gradient and line-search propagation. Equation~\eqref{eq:operator_bound} gives the conservative half-objective Lipschitz bound $L=(c_{\max}/c_{\min})^2+\lambda$.

For numerical accuracy references, we compute solutions of the same constrained objective in double precision. The certificate
\begin{equation}
 \frac{J_c(u;y)-J_c(u_\star;y)}{2\|y\|_2^2}
 \leq\frac{g(u)^\top u+\|g(u)\|_1}{\|y\|_2^2}\label{eq:gap}
\end{equation}
bounds the normalized half-objective gap and measures convergence to the regularized optimum. Equation~\eqref{eq:error} separately measures terminal-field reconstruction.

\section{Experimental Design}\label{sec:experiments}
\subsection{Material and Target Distributions}\label{sec:data}
We use layered and inclusion media to test inverse design with extended interfaces and localized variations in wave speed. Layered media have an inclined interface separating speeds sampled from $[0.70,0.95]$ and $[1.20,1.50]$. Inclusion media contain three ellipses with speeds in $[0.65,1.50]$ in a unit-speed background. Interface positions, slopes, and inclusion geometries vary across samples; Supplementary Material S1 gives their sampling distributions.

Each data split balances two target classes that distinguish a prescribed shape from a field known to be attainable. Geometric targets are nonnegative patterns containing one to five spots, disks, or line segments. Gaussian smoothing and a boundary taper produce fields with peak amplitude at most one. These targets specify desired shapes independently of the propagation operator; their geometric sampling and smoothing parameters are given in Supplementary Material S1.

Reachable targets are obtained by propagating admissible initial controls
\begin{equation}
 u_{\mathrm{gen}}(x,y)=s\sum_{k,l=1}^{8}a_{kl}\sin(k\pi x)\sin(l\pi y),\label{eq:witness}
\end{equation}
where independent Gaussian coefficients have standard deviation $1/(k^2+l^2)$ and $s$ scales the peak nodal amplitude to a value sampled from $[0.25,0.8]$. The generating control establishes discrete reachability, but only the resulting terminal target and speed map enter training.

The compact study uses 16,384 pairs sampled from 4,096 material maps. Its validation set has 256 pairs from a separate 128-map pool. Two original test splits each contain 1,024 new targets, on either encountered training materials or an independent pool of 256 material maps. Capacity screening expands training to 65,536 pairs from the same training-material pool while preserving the original pairs. Target arrays are distinct across training, validation, and test splits.

\subsection{Network and Training Settings}\label{sec:implementation}
We use $n=64$ interior nodes per direction, terminal time $T=0.2$, and $M=160$ time steps. For speeds in $[0.65,1.5]$, the largest Courant number is approximately 0.122. The propagator is implemented as a compiled scan in JAX \cite{jax2018github}; Flax and Optax provide the network and optimizer implementations.

Each U-Net block contains two $3\times3$ convolutions with Gaussian error linear unit (GELU) activations. The encoder widths are $w,2w,4w$, with average pooling between scales and a width-$4w$ bottleneck. Bilinear upsampling and concatenation with encoder features feed decoder blocks of widths $4w,2w,w$. A $1\times1$ convolution and $\tanh$ produce the initial pressure. The width-16 and width-64 models contain 302,705 and 4,832,705 parameters, respectively.

We set $\lambda=2\times10^{-4}$ and train each main model for 60,000 Adam updates with batch size 16 and cosine learning-rate decay from $3\times10^{-4}$ to $3\times10^{-5}$. Validation is performed every 1,000 updates. The selected checkpoint minimizes validation mean squared relative field error, without the control penalty. Computational costs are measured on NVIDIA A40 GPUs; training cost refers to synchronized optimizer updates. Supplementary Material S1 gives the software configuration and measurement procedures.

\subsection{Accuracy Assessment and Model Comparisons}\label{sec:evaluation}
We assess a predicted control by the field it produces at the design time. Using the same discrete propagator for every control, the relative terminal-field error is
\begin{equation}
 e(y,c)=\frac{\|A_cF_\theta(y,c)-y\|_2}{\|y\|_2}.\label{eq:error}
\end{equation}
An independent NumPy implementation evaluates propagation in float64. We average $e$ separately for each target class and material split, so differences in target difficulty remain visible. For models trained with three seeds, we report the mean and sample standard deviation of their mean errors.

The model comparisons distinguish the effects of capacity, training coverage, and sign symmetry. A compact study compares ordinary and wider U-Nets, random sign augmentation, the shared sign projection, and material-modulated architectures based on feature-wise conditioning \cite{perez2018film}. Width and data coverage are selected on the original validation set. At larger capacity, widths 64, 96, and 128 test whether further enlargement improves reconstruction with fixed training data and update budget (Table~\ref{tab:larger}). Supplementary Material S1 specifies the architectures and training comparisons.

We assess the selected width-64 model and compact baseline on an independent 2,048-case confirmation set, using three training seeds. Each known/unseen material split contains 512 targets of each class, with unseen materials drawn from a fresh 256-map pool. At width 64, an additional paired analysis isolates the effect of the sign projection by matching initialization, minibatch sequences, and learning-rate schedules. It reuses the confirmation cases without further model selection (Table~\ref{tab:paired}).

Numerical inversion provides a second point of comparison by directly optimizing the control for the same constrained wave model. We use 64 cases, selected before prediction and balanced across material splits, target classes, and material families. Their reference controls satisfy the normalized half-objective gap tolerance of $10^{-5}$ in Equation~\eqref{eq:gap}. Comparing their field errors with the neural predictions separates the residual of the regularized inverse problem from the remaining error in the learned map.

We also examine how reconstruction responds to changes in the physical setting. Smooth speed maps and six-inclusion media test changes in material structure, with 64 targets per class and family. Varying $\lambda$ over $2\times10^{-5}$, $2\times10^{-4}$, and $2\times10^{-3}$ tests the balance between pressure amplitude and field fidelity. Grid refinement tests whether a designed control retains its reconstruction accuracy under finer propagation. Each coarse-grid control is extended piecewise bilinearly through the zero boundary, while material interfaces are reconstructed from their geometry. We propagate it on grids with 64, 129, and 259 interior nodes and 160, 320, and 640 time steps, comparing errors at shared coarse-grid nodes.

\subsection{Inverse Design under an Online Time Budget}\label{sec:budget_protocol}
A reusable inverse map is useful when a new design is needed before an iterative solver has time to converge. We test this setting by comparing direct prediction with projected gradient, FISTA, and restarted FISTA under a primary 5\,ms median online budget. Secondary budgets of 10 and 20\,ms show how the comparison changes as numerical inversion has more time. The benchmark uses smooth geometric targets and reachable targets generated from $2\times2$ or $8\times8$ sine-mode controls, with speed contrasts $r\in\{1,1.5,2.1\}$. This distribution follows an 80-case development study; its sampling rules, network checkpoints, solvers, and budgets were fixed before generating the independent benchmark. Supplementary Material S1 gives the target and material construction.

Pairing the cases across target structure and speed contrast lets us compare methods on the same underlying material geometries. Sixteen independent descriptors, eight per material family, produce 144 cases across nine target/contrast conditions. The primary comparison uses the 96 heterogeneous cases with equal condition weights. The 48 homogeneous controls help interpret convergence through the wave modes of a simpler medium. Neural results average the errors of three separately evaluated models. Paired 95\% confidence intervals resample descriptor IDs within material families, keeping all conditions associated with each descriptor together.

Each numerical query starts independently from $u_0=0$, with momentum and line-search state reset. For each solver and target/contrast condition, we select the largest measured iteration count whose median A40 latency fits the budget, using timing alone. The budget measures control generation after compilation, with inputs already on the device; offline training, transfers, and final field verification are outside this interval. All resulting controls are evaluated using the same float64 field-error metric defined above. Warmup, synchronization, repeated measurements, and bootstrap settings are specified in Supplementary Material S1.

\section{Numerical Results}\label{sec:results}
\subsection{Reconstruction across Targets and Material Realizations}\label{sec:capacity}
The clearest distinction in the confirmation results is between target classes, rather than between known and unseen materials (Table~\ref{tab:confirmation}). The width-64 model reaches 25.24\% error on unseen geometric targets and 4.77\% on unseen reachable targets, with little change from the corresponding known-material errors. Within the tested families, the learned map therefore accommodates new speed maps more readily than it reproduces an independently prescribed geometric shape.

A larger model trained on more target pairs improves both classes in every seed. Relative to the compact configuration, unseen geometric and reachable errors fall by 17.6\% and 42.0\%, respectively, and the upper tail of the error distribution also contracts (Supplementary Material S1). The gain thus extends beyond the easiest cases, although it costs approximately 2.7 times as much per optimizer update on A40. The much larger error remaining for geometric targets motivates the capacity and accuracy-reference comparisons below.
\begin{table}[H]
\centering\small
\caption{Independent confirmation: mean and sample standard deviation across three training seeds, each with 60,000 updates, with 512 cases per split/class. Both models are symmetrized; errors are percentages. Costs on A40 are mean time per Adam update and per-model inference latency medians; batch-16 time covers the whole batch. Measurement procedures are given in Supplementary Material S1.}
\label{tab:confirmation}
\setlength{\tabcolsep}{4pt}
\begin{tabular}{lrrrr}
\toprule
& \multicolumn{2}{c}{Known materials} & \multicolumn{2}{c}{Unseen materials}\\
Model & Geometry & Reachable & Geometry & Reachable\\
\midrule
Compact ($w=16$) & $30.77\pm0.048$ & $8.17\pm0.005$ & $30.64\pm0.066$ & $8.23\pm0.042$ \\
Selected ($w=64$) & $25.32\pm0.039$ & $4.69\pm0.023$ & $25.24\pm0.050$ & $4.77\pm0.023$ \\
\bottomrule
\end{tabular}
\smallskip

\begin{tabular}{lrrr}
\toprule
Model & Update (ms) & Network batch 1 (ms) & Network batch 16 (ms)\\
\midrule
Compact ($w=16$) & $14.267\pm0.145$ & $0.671\pm0.001$ & $1.497\pm0.002$ \\
Selected ($w=64$) & $38.035\pm0.008$ & $4.259\pm0.002$ & $12.779\pm0.006$ \\
\bottomrule
\end{tabular}
\end{table}

\begin{figure}[H]
\centering\includegraphics[width=\textwidth]{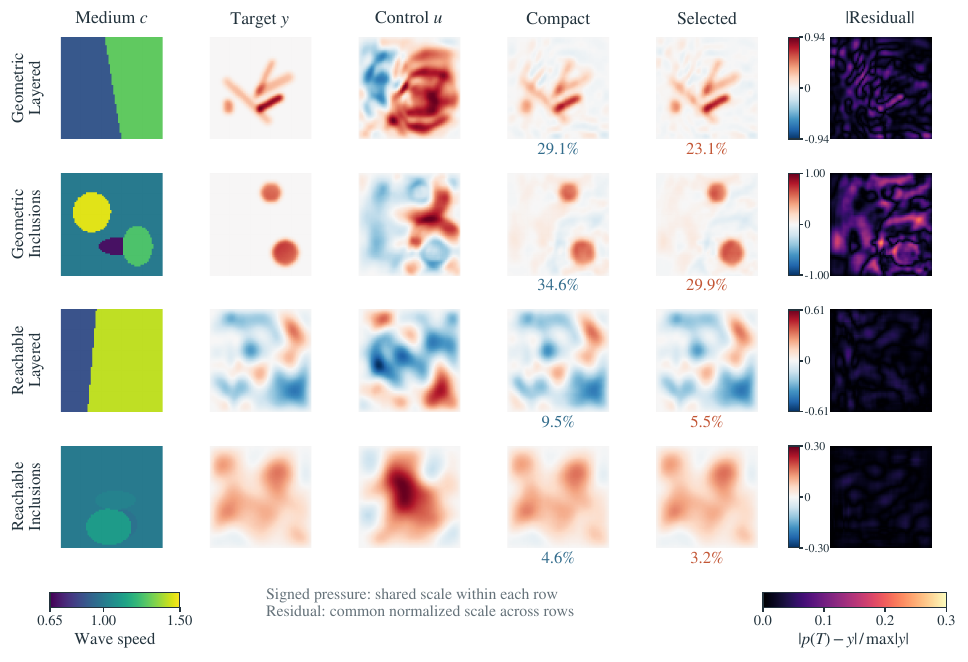}
\caption{Preselected unseen confirmation cases, one per target/material combination. Columns show speed, target, selected initial pressure, compact and selected terminal fields, and selected absolute residual. Both models are symmetrized; seed-zero relative errors are percentages. Signed pressure shares one scale per row. Residuals are normalized by each target's peak absolute amplitude and use a separate common scale; speeds range from 0.65 to 1.5.}
\label{fig:examples}
\end{figure}

The predicted controls in Figure~\ref{fig:examples} reveal why target shape alone does not describe the design problem. For the nonnegative geometric targets shown here, the network predicts signed initial patterns whose positive and negative contributions propagate and interfere at the prescribed time. The selected model reduces terminal-field error in each displayed case, but the residuals show that reproducing this interference pattern remains imperfect.

\subsection{Effect of the Sign-Symmetric Parameterization}\label{sec:symmetry_results}
The capacity comparison changes both model size and training coverage. At fixed width, the paired comparison identifies a separate gain from the sign-symmetric parameterization: unseen geometric and reachable errors fall by 9.0\% and 28.6\%, with improvements in all three seeds (Table~\ref{tab:paired}). The number of trainable parameters is unchanged. Enforcing the pressure-reversal relation therefore improves reconstruction empirically, including for nonnegative geometric targets.
\begin{table}[H]
\centering\small
\caption{Paired comparison of the sign parameterization: both models have 4,832,705 parameters, 65,536 training pairs and 60,000 updates. Mean and sample standard deviation across three seeds; 512 cases per split/class. This additional paired analysis reuses the previously evaluated confirmation cases. Update costs use the common A40 measurement protocol in Supplementary Material S1.}
\label{tab:paired}
\setlength{\tabcolsep}{2pt}
\begin{tabular}{lrrrrr}
\toprule
& \multicolumn{2}{c}{Known materials} & \multicolumn{2}{c}{Unseen materials} & Update time\\
Model & Geometry (\%) & Reachable (\%) & Geometry (\%) & Reachable (\%) & (ms)\\
\midrule
Plain ($w=64$) & $27.89\pm0.043$ & $6.58\pm0.055$ & $27.70\pm0.053$ & $6.67\pm0.043$ & $21.83\pm0.04$ \\
Symmetrized ($w=64$) & $25.31\pm0.038$ & $4.68\pm0.020$ & $25.22\pm0.052$ & $4.76\pm0.024$ & $38.07\pm0.17$ \\
\bottomrule
\end{tabular}
\end{table}

Random sign augmentation exposes training to the same physical relation but leaves the network free to violate it. In the compact seed-zero comparison, the exact projection yields lower errors than this augmentation for both target classes on known and unseen materials (Supplementary Material S1, Table S4). This comparison favors building the relation into the prediction for the present architecture.

\begin{figure}[H]
\centering\includegraphics[width=\textwidth]{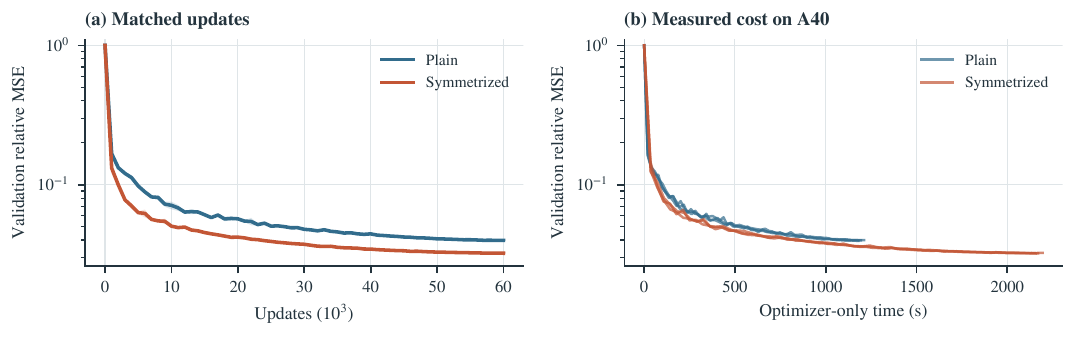}
\caption{Paired validation mean squared relative field error for width-64 models. The update panel reports the mean and sample standard deviation across three seeds, including initialization. The time panel shows each run against its synchronized A40 optimizer time, defined in Section~\ref{sec:implementation}.}
\label{fig:validation}
\end{figure}

The projection requires two network evaluations, increasing the cost per optimizer update by a factor of approximately 1.7 (Table~\ref{tab:paired}). Figure~\ref{fig:validation} shows that its reconstruction benefit is still visible when training is compared by optimizer time: within the plain models' training-time budgets, symmetrization reaches a 9.1\% lower mean best recorded validation squared relative error. Both models retain their update-based learning-rate schedules; Supplementary Material S1 gives the corresponding time limits and update counts.

\subsection{Capacity, Data Coverage, and Training Duration}\label{sec:scaling}
To understand the improvement from the compact to the selected model, we separate changes in width, data coverage, and training duration. At compact scale, width has the larger effect: doubling it from 16 to 32 reduces the selection metric by 16.9\%, whereas quadrupling the data at width 16 reduces it by only 0.8\%. The narrow network appears unable to make much use of the additional pairs. Expanding both width and data gives a larger gain, while extending width-48 training brings only a small further improvement (Supplementary Material S1).

This trend does not continue at the same rate beyond width 64 (Table~\ref{tab:larger}). Width 128 lowers geometric error only from 24.22\% to 23.42\%, despite approximately four times as many parameters and 2.5 times the A40 cost per optimizer update. Reachable targets benefit more in relative terms. Width 64 is consequently a useful balance for this training schedule, but the persistent geometric error calls for a closer look at the inverse problem itself.

\begin{table}[H]
\caption{Exploratory expansion beyond width 64 on the original 256 validation cases. Each model is trained from scratch with seed zero, 65,536 pairs, and 60,000 updates. Errors are means of relative terminal-field errors; $\overline{e^2}$ selects checkpoints. A40 update costs follow the measurement procedure in Supplementary Material S1.}
\label{tab:larger}
\centering\small
\begin{tabular}{rrrrrr}
\toprule
$w$ & Parameters & $\overline{e^2}$ & Geometry (\%) & Reachable (\%) & Update (ms)\\
\midrule
64 & 4,832,705 & 0.03219 & 24.22 & 4.59 & 38.01\\
96 & 10,870,945 & 0.03063 & 23.66 & 4.13 & 66.86\\
128 & 19,323,777 & 0.02997 & 23.42 & 3.96 & 93.96\\
\bottomrule
\end{tabular}
\end{table}

\subsection{Target Structure and the Numerical Accuracy Reference}\label{sec:target_accuracy}
Direct numerical inversion helps distinguish errors associated with the target from those remaining in the learned map. On the matched unseen reference cases, the regularized numerical solution reaches 0.21\% error for reachable targets, compared with 4.30\% for the selected network. Geometric targets retain much larger errors in both methods: 14.15\% and 25.37\%, respectively. Unlike a reachable target, a geometric shape is specified independently of propagation and the pressure bound. Its larger numerical residual therefore reflects the difficulty of reproducing that shape under the regularized design objective, while the additional neural error leaves room to improve the learned inverse.

The spatial content of the targets helps explain this distinction. Reachable controls emphasize low-order sine components, whereas geometric shapes contain finer features: 9.18\% of their energy lies outside the first $8\times8$ sine modes on the original test cases, compared with 0.30\% for reachable targets. Producing these features requires pressure contributions that compensate for the medium's response at the corresponding scales. This spectral contrast identifies one aspect of the harder design task; the numerical residual also includes the trade-off between field fidelity and control amplitude.

\subsection{Sensitivity to Regularization, Material Families, and Propagation Resolution}\label{sec:robustness}
Because the numerical reference balances amplitude and fidelity, the control penalty is a natural place to look for a change in reconstruction. In the compact-model study, however, a hundredfold increase in $\lambda$ lowers geometric-control root mean square (RMS) by about 5\% while leaving field error close to 29.5\%. The network's reachable-target errors also change little, even though the numerical inverse responds more strongly to the penalty. Adjusting this trade-off alone therefore offers little improvement in neural reconstruction over the tested range. Supplementary Material S1 gives the complete comparison.

Changing material structure has a stronger effect than changing the realization within a familiar family. Both smooth speed maps and six-inclusion media fall outside the training families. The selected model improves on the compact one in both probes, but smooth media produce larger errors (Table~\ref{tab:ood}). Smoothness by itself therefore does not make an unfamiliar inverse response easier to predict for a network trained on layers and sharp inclusions. Generalization depends on the spatial organization of the medium as well as its speed range.
\begin{table}[H]
\centering\small
\caption{Transfer of the confirmation models to smooth and six-inclusion media. Errors (\%) report three-seed mean and sample standard deviation; each family has 64 geometric and 64 reachable cases. No adaptation or selection uses these cases.}
\label{tab:ood}
\setlength{\tabcolsep}{4pt}
\begin{tabular}{lrrrr}
\toprule
& \multicolumn{2}{c}{Smooth media} & \multicolumn{2}{c}{Six inclusions}\\
Model & Geometry & Reachable & Geometry & Reachable\\
\midrule
Compact ($w=16$) & $38.92\pm0.200$ & $13.60\pm0.096$ & $31.72\pm0.112$ & $9.79\pm0.058$ \\
Selected ($w=64$) & $35.17\pm0.187$ & $9.63\pm0.166$ & $26.69\pm0.028$ & $6.02\pm0.076$ \\
\bottomrule
\end{tabular}
\end{table}

\begin{figure}[H]
\centering\includegraphics[width=\textwidth]{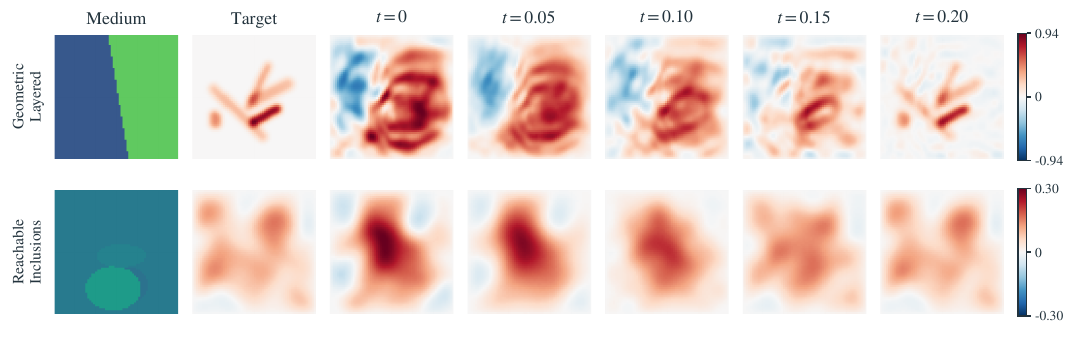}
\caption{Free evolution of two width-64 controls for unseen geometric/layered and reachable/inclusion cases, selected independently of prediction quality. Each row uses one pressure scale; speed ranges from 0.65 to 1.5.}
\label{fig:trajectory}
\end{figure}

The propagation grid introduces another kind of change: the intended medium and target remain the same, but the discrete wave response becomes finer. Figure~\ref{fig:trajectory} follows the controls through time; Table~\ref{tab:refinement} then holds those controls fixed and refines the propagation grid. The selected model's geometric error changes only from 25.37\% to 25.89\%, with nearly unchanged reachable-target error. Interpolated coarse-grid numerical controls lose more geometric accuracy, showing how a design can depend on details of its discrete propagator. For reachable targets, numerical errors remain below the neural errors at every tested resolution.

\begin{table}[H]
\centering\small
\caption{Fixed-control refinement on the same unseen reference cases, 16 per target class. Neural errors (\%) report mean and sample standard deviation across the three frozen confirmation seeds. Errors are evaluated at shared coarse-grid nodes. Numerical controls are independently certified on the coarse grid and then held fixed.}
\label{tab:refinement}
\setlength{\tabcolsep}{5pt}
\begin{tabular}{llrrr}
\toprule
Control & Target & $n=64$ & $n=129$ & $n=259$\\
\midrule
Compact & Geometry & $30.58\pm0.050$ & $30.79\pm0.043$ & $30.83\pm0.045$ \\
Selected & Geometry & $25.37\pm0.038$ & $25.80\pm0.027$ & $25.89\pm0.035$ \\
Numerical & Geometry & 14.15 & 29.18 & 31.16 \\
\midrule
Compact & Reachable & $7.17\pm0.112$ & $7.14\pm0.108$ & $7.14\pm0.108$ \\
Selected & Reachable & $4.30\pm0.056$ & $4.32\pm0.058$ & $4.34\pm0.056$ \\
Numerical & Reachable & 0.21 & 0.94 & 1.05 \\
\bottomrule
\end{tabular}
\end{table}

Reoptimizing the control on the finer grid recovers much of the lost accuracy, reaching 4.68\% mean common-node error on eight stratified cases. This recovery ties the deterioration to transferring the coarse-grid design, rather than to refinement alone. The finer-grid solves reached their iteration cap before the specified optimality threshold (Supplementary Material S1).

\subsection{Reconstruction under Online Computation Budgets}\label{sec:budget_results}
The accuracy references show what numerical inversion can recover after substantial iteration. The online benchmark instead asks what control is available within a few milliseconds. On the smooth geometric and low- to mid-frequency reachable targets defined in Section~\ref{sec:budget_protocol}, JAX-WAVE reaches 18.79\% mean field error within the primary 5\,ms median budget, compared with 34.18\% for FISTA (Table~\ref{tab:budget}). The pooled relative reduction is 45.0\%, with a paired 95\% interval of [40.4\%,49.4\%]. Both target classes improve, and each network seed retains a pooled advantage.
\begin{table}[H]
\centering\small
\caption{Independent heterogeneous-media benchmark within a 5\,ms median online computation budget on A40. Lower mean terminal-field error is better. The 32 geometric and 64 reachable cases comprise six equally weighted conditions. Neural errors average three separately evaluated seeds; numerical budgets use timing-only iteration selection. The final column reports relative error reduction by JAX-WAVE against each baseline, with paired 95\% bootstrap confidence intervals (CI).}
\label{tab:budget}
\setlength{\tabcolsep}{5pt}
\begin{tabular}{lrrrr}
\toprule
Method & Geometry (\%) & Reachable (\%) & Pooled (\%) & Reduction (\%) [95\% CI]\\
\midrule
Projected gradient & 45.65 & 31.94 & 36.51 & 48.5 [44.3,52.6]\\
FISTA & 44.03 & 29.25 & 34.18 & 45.0 [40.4,49.4]\\
Restarted FISTA & 44.03 & 29.25 & 34.18 & 45.0 [40.4,49.4]\\
JAX-WAVE & \textbf{31.87} & \textbf{12.25} & \textbf{18.79} & ---\\
\bottomrule
\end{tabular}
\end{table}

\begin{figure}[H]
\centering\includegraphics[width=\textwidth]{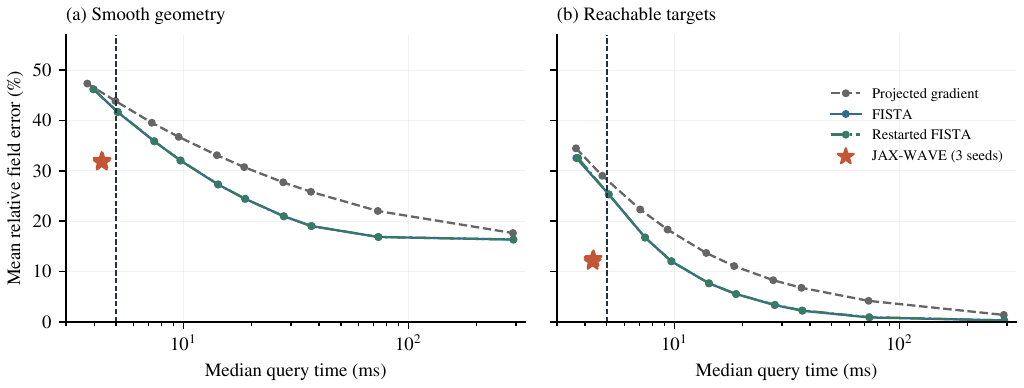}
\caption{Accuracy versus control-generation time on the independent heterogeneous-media benchmark. Panels separate smooth geometric and reachable targets. Numerical curves use uniform iteration counts; stars show the three frozen networks. The dashed line marks 5\,ms. Table~\ref{tab:budget} selects iteration counts per condition using timing alone. Measurements use synchronized A40 execution under the control-generation timing protocol in Section~\ref{sec:budget_protocol}, which excludes terminal-field verification.}
\label{fig:budget_latency}
\end{figure}

The network generates a control in a median 4.32\,ms, while FISTA can complete only three or four updates within the primary budget. More iterations change the balance (Figure~\ref{fig:budget_latency}): FISTA reaches 18.71\% pooled error at 10\,ms, comparable to the network, and improves to 11.84\% at 20\,ms. The learned map makes information acquired during training available immediately for a new query; numerical inversion improves its answer as computation proceeds. Their relative value depends on how soon a control is needed and how much reconstruction error the application can accept.

To understand why even smooth targets can be slow to recover iteratively, we examine the homogeneous-medium controls. Each sine mode evolves with its own temporal factor, so its terminal amplitude depends on the chosen observation time as well as its spatial frequency. At the terminal time used here, several low modes have weak transfer and small objective curvature (Appendix~\ref{app:modal}), making them slow directions for a global gradient step. On the 16 homogeneous $2\times2$-mode cases, four-step FISTA gives 38.69\% mean field error, compared with 2.33--2.80\% across the three trained networks. The $(1,2)$, $(2,1)$, and $(2,2)$ modes account for 94.2\% of the FISTA residual energy, locating most of the remaining mismatch in these weak-response directions. The diagnostic connects early reconstruction error to the wave response at the design time; the direct benchmark measures the corresponding performance comparison in heterogeneous media.

\subsection{Repeated-Query Cost}\label{sec:amortization}
The latency advantage is available after training; repeated use determines whether those per-query savings also repay the training cost. The eight-case A40 comparison in Supplementary Material S1 provides an estimate at similar reconstruction accuracy: the paired symmetrized model gives 12.09\% mean error in 3.265\,ms, while 16-step backtracking FISTA gives 12.18\% in 18.662\,ms. Using the mean A40 optimizer time of 2,173.1\,s for one training run, the query count that offsets this cost is
\begin{equation}
 N_\star=\frac{C_{\mathrm{train}}}{t_{\mathrm{num}}-t_{\mathrm{net}}}\approx1.41\times10^5.\label{eq:amortization}
\end{equation}
For this eight-case workload, the measured training-update cost is recovered after roughly 141,000 queries. Data preparation, compilation, and model selection add to that offline investment; Supplementary Material S1 specifies the measurement protocol. Amortization is therefore most relevant to sustained reuse across design queries. When queries repeatedly use one material, cached numerical operators provide another way to reduce their cost, as the same supplement shows.

\section{Discussion}\label{sec:discussion}
\subsection{Learning across Targets and Media}
Differentiable physical models support inverse learning by linking unknown fields to their observable response, including displacement and material fields in hyperelastic analysis \cite{du2025hyper}. For repeated wave design, we learn how the required initial pressure changes with both the target and the medium. Spatial variations in wave speed change how pressure contributions propagate and combine at the design time. Conditioning on the speed map gives the network this environmental information, and propagation-based training tests the resulting control through the wave equation. The similar errors on known and unseen realizations show that the learned response transfers within the tested material families. The smooth-media probes expose a more demanding change in spatial structure.

Pressure-sign symmetry supplies a simple physical organizing principle: reversing compression and rarefaction reverses the wave field while leaving the medium unchanged. The shared projection builds this relation into every prediction. Its paired-objective identity describes a fixed network, and the training comparisons establish the reconstruction benefit, including for nonnegative geometric targets. The improvement at matched optimizer time makes this constraint useful beyond its exact algebraic property. This use of known structure as an architectural constraint has parallels in neural causal models \cite{xia2023ncm}.

\subsection{Target Structure and Approximation Accuracy}
The numerical reference separates two questions: how accurately the constrained wave system reproduces a prescribed shape, and how closely the network approximates the corresponding inverse design. Reachable targets are nearly recovered by numerical inversion, leaving a clear approximation gap for the network. Geometric targets retain larger residuals in both approaches. The two target classes differ in both construction and spatial content, and the control penalty further shapes the solution. A sharper requested image need not correspond to an easily generated wave field.

The limited gain from widening the network suggests that the next improvement should address how the inverse response is represented. More capacity helps, but it does not by itself resolve the geometric-target errors under the tested training budget.

Studies of spectral bias report frequency-dependent learning in neural networks \cite{rahaman2019spectral}, and Fourier features improve high-frequency fitting in coordinate-based multilayer perceptrons \cite{tancik2020fourier}. Frequency and scale dependence have also motivated transfer-learning strategies for residual-based physics-informed networks \cite{mustajab2024transfer}. These studies motivate testing frequency-aware representations for the present inverse map. Such a representation would need to account for both the spatial content of the target and the medium-dependent propagation response.

\subsection{Practical Use and Scope}
Direct prediction is useful when many design queries share the trained physical setting and online computation is limited. The 5\,ms benchmark demonstrates this use case on smooth geometric and low- to mid-frequency reachable targets at moderate contrasts. The 10 and 20\,ms results locate the crossover to numerical inversion on that workload. The development sweep also contains fine geometric targets, higher-frequency controls, and contrast 3.0 cases that often favor numerical iteration. Offline training and data generation remain part of the cost of preparing a shared model; the online measurements quantify the control-generation stage after training. The amortization estimate in Section~\ref{sec:amortization} makes the repeated-query requirement explicit: low online latency becomes a saving in total computation only after sufficient reuse.

The present map shares a fixed grid, terminal time, boundary condition, and initial pressure derivative across queries. Making the terminal time an input would extend design from a single prescribed instant to a family of propagation horizons. Because the response of each wave mode changes with time, this extension would require training across those horizons.

A route toward physical deployment is to replace distributed pressure control by calibrated actuator responses. Writing $u=Ba$ maps actuator amplitudes $a$ through spatial responses $B$, connecting the current formulation to source-plane or transducer-array designs \cite{diffpat2021,lee2022}. A network could then predict $a$ and train through the resulting propagation while enforcing actuator limits. The sign projection remains feasible when the admissible amplitude set is convex and sign-symmetric.

\section{Conclusions}\label{sec:conclusion}
JAX-WAVE formulates repeated wave-field design as a physics-supervised, amortized inverse problem. A shared model maps desired terminal fields and heterogeneous speed maps to bounded initial pressures. Differentiable propagation supplies the training signal, and pressure-sign symmetry incorporates an exact relation of the physical inverse problem into the prediction. The discrete operator analysis connects the same propagation model to its adjoint and numerical accuracy reference, while paired comparisons show that the sign constraint improves reconstruction without enlarging the trainable model.

The most informative distinction in the experiments is between target classes. Transfer to new material realizations causes little change within the tested families, while geometrically prescribed shapes remain much harder to reproduce than targets generated by admissible controls. Numerical references expose a remaining gap in the learned inverse map, and wider networks yield diminishing returns. These findings motivate representations that better capture how target structure interacts with propagation. For repeated design queries, the trained map already provides bounded controls within a short online budget, making the wave physics learned offline available at the time a new design is requested.

\section*{Author Contributions}
Honghui Du conceived the study and contributed to the numerical experiments and manuscript writing. Binyao Guo and Yushu Pan contributed to the numerical experiments and manuscript writing.
\section*{Acknowledgments}
Computations used the Minnesota Supercomputing Institute.
\section*{Conflict of Interest}
The authors declare no conflict of interest.
\section*{Funding}
This research received no external funding.
\section*{Data Availability Statements}
The code, reported-study summaries, numerical checks, and independent time-budget benchmark are available at \url{https://github.com/Honghui-Du/JAX-WAVE}. The repository links the original reproduction archive containing datasets, frozen checkpoints, training histories, and optimizer states. The additional width-screen records accompany the manuscript as Supplementary Data S2.
\section*{Supplementary Material}
Supplementary Material S1 gives detailed experimental protocols and additional results, including Figures S1--S4 and Tables S1--S9. Supplementary Data S2 contains machine-readable records for the completed width-64/96/128 validation screen.

\appendix
\section{Discrete Propagation, Adjoint, and Optimality Certificate}\label{app:operator}
Write $B=C^2D_h$ and $\tau=\Delta t$. The Taylor start and centered recurrence imply
\begin{equation}
 A_c=T_M(I+\tau^2B/2),
\end{equation}
where $T_M$ is the degree-$M$ Chebyshev polynomial of the first kind. With $S=CD_hC$, the similarity $B=CSC^{-1}$ yields
\begin{equation}
 A_c=CPC^{-1},\qquad P=T_M(I+\tau^2S/2)=P^\top.
\end{equation}
Consequently $A_c^\top=C^{-1}PC=C^{-2}A_cC^2$, giving Equation~\eqref{eq:adjoint}. Under $\tau c_{\max}/h\leq1/\sqrt2$, the spectrum of $I+\tau^2S/2$ lies in $[-1,1]$, so $\|P\|_2\leq1$. Therefore
\begin{equation}
 \|A_c\|_2\leq c_{\max}/c_{\min},\qquad
 L=(c_{\max}/c_{\min})^2+\lambda
\end{equation}
is a conservative Lipschitz bound for the gradient of $J_c/2$.

For an admissible $u$, convexity and the box constraint give
\begin{equation}
 \tfrac12[J_c(u;y)-J_c(u_\star;y)]
 \leq g(u)^\top(u-u_\star)
 \leq g(u)^\top u+\|g(u)\|_1.
\end{equation}
Dividing by $\|y\|_2^2$ proves Equation~\eqref{eq:gap}. Strong convexity and constrained optimality also give $J_c(u;y)-J_c(u_\star;y)\geq\lambda\|u-u_\star\|_2^2$. A normalized half-objective gap of at most $\varepsilon$ thus implies
\begin{equation}
 \frac{\|u-u_\star\|_2}{\|y\|_2}\leq\sqrt{\frac{2\varepsilon}{\lambda}}.
\end{equation}
This bound is normalized by the target norm, not the unknown optimal-control norm.

\section{Homogeneous Modal Diagnostic}\label{app:modal}
For $c=1$, the discrete Dirichlet sine modes, indexed by $k,l=1,\ldots,n$, have squared frequencies
\begin{equation}
 \omega_{kl}^2=4(n+1)^2\left[\sin^2\!\left(\frac{k\pi}{2(n+1)}\right)+\sin^2\!\left(\frac{l\pi}{2(n+1)}\right)\right].
\end{equation}
The terminal transfer and half-objective curvature are
\begin{equation}
 a_{kl}=\cos\!\left[M\arccos\!\left(1-\frac{\tau^2\omega_{kl}^2}{2}\right)\right],
 \qquad q_{kl}=a_{kl}^2+\lambda.
\end{equation}
For the experimental discretization, $a_{11}=0.63058$, $a_{12}=a_{21}=0.16553$, and $a_{22}=-0.20423$. The corresponding curvatures are 0.39784, 0.02760, 0.02760, and 0.04191, and the condition number of the regularized normal matrix $A_c^\top A_c+\lambda I$ is approximately 4,926.6. An unconstrained gradient step of size $\alpha$ multiplies each modal control error by $1-\alpha q_{kl}$, explaining why weakly curved directions progress slowly under a global step. Projection and acceleration alter this scalar iteration, so the calculation diagnoses the objective rather than proving a convergence rate for constrained FISTA. The residual-energy analysis in Section~\ref{sec:budget_results} uses an orthonormal type-I discrete sine transform and sums squared coefficients over the stated cases and modes.

\bibliographystyle{unsrtnat}
\bibliography{references}
\clearpage
\setcounter{section}{0}
\setcounter{subsection}{0}
\setcounter{equation}{0}
\setcounter{table}{0}
\setcounter{figure}{0}
\renewcommand{\thesection}{S\arabic{section}}
\renewcommand{\thetable}{S\arabic{table}}
\renewcommand{\thefigure}{S\arabic{figure}}
\renewcommand{\theequation}{S\arabic{equation}}
\renewcommand{\theHsection}{supp.\arabic{section}}
\renewcommand{\theHsubsection}{supp.\arabic{section}.\arabic{subsection}}
\renewcommand{\theHequation}{supp.\arabic{equation}}
\renewcommand{\theHtable}{supp.\arabic{table}}
\renewcommand{\theHfigure}{supp.\arabic{figure}}
\begin{center}
{\Large\bfseries Supplementary Material S1}
\end{center}

This supplement describes the experimental protocols and additional results.

\section{Data, Models and Evaluation}
\subsection{Computational Environment}
All reported timing comparisons use NVIDIA A40 GPUs. The confirmation and paired studies use JAX 0.4.30, Flax 0.8.5, and Optax 0.2.4 with highest-precision float32 matrix/convolution settings and deterministic GPU operations. The original compact-model training uses JAX 0.10.2, Flax 0.12.8, and Optax 0.2.8. Deployment costs use the A40 software configuration.

For the model comparisons (main Tables 1--3), update cost averages 100 synchronized calls at the saved parameter state with freshly initialized Adam moments and batch size 16, after compilation and ten warmups. Parameters and optimizer states are held fixed between timing calls. Inference compiles the fixed parameters into the prediction function and uses 20 warmups and 100 synchronized repeats with resident inputs.

Optimizer time excludes compilation, input transfer, validation, and checkpointing.

\subsection{Materials and Targets}
Two material families cover sharp changes in wave speed. Layered media use the interface $x=a+b(y-0.5)$, with $a\sim\mathcal U(0.25,0.65)$ and $b\sim\mathcal U(-0.25,0.25)$. Speeds on the two sides are sampled from $[0.70,0.95]$ and $[1.20,1.50]$. Inclusion media have unit background speed and three elliptical inclusions. Their centers are sampled from $[0.2,0.8]^2$, their semiaxes from $[0.08,0.22]$, and their speeds from $[0.65,1.50]$. Later inclusions overwrite earlier ones where they overlap.

Each split contains equal numbers of two target classes. \emph{Geometric targets} are nonnegative patterns comprising one to five spots, disks or line segments. Each target uses one primitive type, sampled with probabilities 0.2/0.4/0.4, and primitive amplitudes sampled uniformly from $[0.3,1]$. These patterns test direct specification of a desired terminal shape. Object centers and line endpoints are sampled from $[0.15,0.85]^2$, with disk/spot radius parameters in $[0.06,0.15]$ and line widths in $[0.06,0.12]$. Gaussian smoothing has physical standard deviation $0.7/33$, followed by the boundary taper $[\sin(\pi x)\sin(\pi y)]^{1/4}$ and amplitude normalization to a maximum of at most one. \emph{Reachable targets} are generated by propagating initial pressures
\begin{equation}
 u_{\rm gen}(x,y)=s\sum_{k,l=1}^{8}a_{kl}\sin(k\pi x)\sin(l\pi y),\label{supp:eq:witness}
\end{equation}
where independent Gaussian coefficients have standard deviation $1/(k^2+l^2)$ and $s$ sets the maximum nodal amplitude to a value sampled from $[0.25,0.8]$. These targets provide a known admissible control that reproduces the requested field; the regularized optimum can still trade field accuracy for control amplitude. Only their terminal fields and material maps are provided to training.

The compact-model study uses 16,384 pairs sampled from a pool of 4,096 material maps, equally divided between the two families; 4,028 maps appear in the sampled pairs. Validation contains 256 pairs from a separate 128-map pool. Two held-out sets each contain 1,024 new targets: one uses materials actually encountered in training, and the other uses an independent 256-map pool from the same families. All target arrays are distinct across splits. The latter test measures generalization to new realizations of the same two material families.

\subsection{Models and Training}
The base network $G_\theta$ receives the two channels $[y,c-1]$; time is not a network input. Its U-Net encoder  has widths $w,2w,4w$, a width-$4w$ bottleneck and decoder widths $4w,2w,w$. Each block contains two $3\times3$ convolutions with GELU activations. Average pooling downsamples features; bilinear upsampling and concatenated encoder features feed the decoder. A $1\times1$ output convolution precedes $\tanh$. The compact configuration uses $w=16$ and 302,705 parameters; the selected larger configuration uses $w=64$ and 4,832,705 parameters.

We compare six compact architectures using identical training pairs and minibatch sequences. Alongside concatenated inputs, we test spatial material modulation inspired by feature-wise conditioning:
\begin{itemize}
\item \emph{U-Net}: concatenated target and material inputs with the architecture in the main paper.
\item \emph{Wide U-Net}: widths 20/40/80, controlling for the larger capacity of material-modulated networks.
\item \emph{Sign augmentation}: the base U-Net trained with independently sampled target sign reversals.
\item \emph{Symmetrized U-Net}: the shared two-sign projection, sharing the base network weights.
\item \emph{Spatial odd}: a separate material encoder supplies gains $1+0.5\tanh(H(c))$ to bias-free target convolutions, followed by $\tanh$ activations. Pooling, upsampling and skip connections preserve oddness in the target.
\item \emph{Spatial affine}: the same modulated architecture with material-dependent additive shifts, removing exact oddness.
\end{itemize}
The material encoder in the last two models has widths 16/32/64 and a width-64 bottleneck. Two target convolutions per block receive spatial gains through $1\times1$ material projections. The affine version adds a further projection for each shift. The wide, spatial-odd and affine models have 472,701, 477,904 and 508,176 parameters, respectively.

These six models receive 60,000 Adam updates with batch size 16, initialization seed zero and a cosine learning-rate schedule from $3\times10^{-4}$ to $3\times10^{-5}$. Validation is evaluated every 1,000 updates, retaining the checkpoint with the smallest mixed mean squared relative field error. The symmetrized model has the lowest selection metric among these six seed-zero models. Held-out evaluation follows checkpoint selection. We also repeat the U-Net, symmetrized U-Net and spatial-odd model with seeds one and two, changing initialization and minibatch sampling while retaining matched case sequences within each seed. Three-run comparisons report the mean and sample standard deviation of per-run mean errors. Complete six-model validation curves accompany the source archive. Training, inference and held-out propagation use float32; validation selection is checked by an independent NumPy float64 recurrence.

Reconstruction is measured by
\begin{equation}
 e(y,c)=\frac{\|A_cF_\theta(y,c)-y\|_2}{\|y\|_2}.\label{supp:eq:error}
\end{equation}
We report the arithmetic mean of $e$ separately by target class and material split. Paired comparisons for the spatial-odd candidate resample material clusters with all their targets retained, using 2,000 bootstrap draws and percentile intervals. These seed-zero intervals measure case/material sampling uncertainty conditional on the trained models.

\paragraph{Capacity and data.}\label{supp:sec:scaling_protocol}
We retain the shared two-sign projection, the loss, batch size and propagation model. Expanded data preserve the original 16,384 pairs and add targets from the same material pool to reach 65,536. Table~\ref{supp:tab:scaling} compares width and data coverage after 60,000 updates and includes 120,000-update continuations. Continuations for $w=16,32,48$ restore optimizer/sampler state and add 60,000 updates with cosine decay from $3\times10^{-5}$ to $3\times10^{-6}$. Selection uses only the original 256 validation cases.

After selecting width 64 and the expanded dataset on validation, we train both this configuration and the compact baseline afresh with seeds 0/1/2 for 60,000 updates. A separate 2,048-case confirmation set contains 1,024 targets on materials shared by both training sets and 1,024 on a fresh 256-map pool, balanced across target classes and material families. Target arrays are distinct across all splits. The configuration and all six checkpoints are fixed before evaluating this test set.

Confirmation errors use independent float64 propagation of float32 predictions. A40 latency measurements follow Section S1.1 and summarize the three seeds separately from the compact numerical benchmark.

\paragraph{Matched symmetry comparison.}\label{supp:sec:paired_protocol}
To isolate the effect of symmetry at the selected capacity, we compare ordinary and symmetrized U-Net at $w=64$, with 65,536 training pairs and 60,000 updates. Each of three paired seeds uses identical initial parameters, minibatch sampling and learning-rate schedule. The two models run sequentially on the same A40 GPU, with the order reversed for seed one. Checkpoints are selected on validation data, and all six runs are reported. This additional paired analysis reuses the previously evaluated 2,048 confirmation cases, without further model selection.

The paired learning-curve figure in the main paper plots validation error against both updates and synchronized optimizer time. At the end of each plain model's optimizer budget, 1,180--1,211 seconds, the mean minimum logged validation squared relative error is 0.03977 for plain U-Net and 0.03617 for symmetrization. This 9.1\% reduction occurs at 32,000--33,000 symmetrized updates. Both models retain their update-based learning-rate schedules. The numerical-reference, material-family and refinement tests use the compact and selected checkpoints from the confirmation study.

\subsection{Regularization and New Media}\label{supp:sec:sensitivity}
Regularization sensitivity is evaluated at $\lambda\in\{2\times10^{-5},2\times10^{-4},2\times10^{-3}\}$. The symmetrized model is trained separately at each value with seed zero and the same data, update budget and checkpoint-selection rule. Numerical inverses use the matching regularization on the same 64 reference cases. We compare reconstruction error together with nodal control RMS, $\|u\|_2/n$.

A separate exploratory probe changes the material families while retaining the physical problem, speed range and target-generation rules. Smooth media use a random $3\times3$ sine expansion with Gaussian coefficients scaled by $1/(k^2+l^2)$ and independent spatial phases, rescaled to $[0.65,1.5]$. Six-inclusion media use six ellipses with the original parameter distributions. Each family contains 64 new geometric and 64 new reachable targets, each with its own material map. We evaluate all three compact and selected confirmation checkpoints directly on these cases, without adaptation. Errors use independent float64 propagation of float32 control predictions. Table~\ref{supp:tab:ood} compares the compact architectures on these material families.

\subsection{Numerical Inversion and Query Cost}
A bounded numerical inverse provides an accuracy reference for the regularized box objective. We use projected FISTA with the same control bounds, regularization and discrete propagation as the networks. For half of $J_c$, the gradient is
\begin{equation}
 g(u)=A_c^\top(A_cu-y)+\lambda u.
\end{equation}
For this discretization, $C^{-1}(C^2D_h)C=CD_hC$ is symmetric, giving
\begin{equation}
 A_c^\top r=C^{-2}A_c(C^2r).\label{supp:eq:adjoint}
\end{equation}
The adjoint therefore requires one more forward propagation. Under the Courant stability condition, $\|A_c\|_2\leq c_{\max}/c_{\min}$, giving the conservative half-objective Lipschitz bound $L=(c_{\max}/c_{\min})^2+\lambda$. Fixed-step FISTA uses $1/L$ and clips each candidate to $[-1,1]$. Backtracking starts at $L=1$ and doubles it until $\|A_c(u-v)\|_2^2+\lambda\|u-v\|_2^2\leq L\|u-v\|_2^2$, where $v$ is the extrapolated iterate and $u$ the projected candidate. The accepted bound carries forward. Appendix A of the main article derives the adjoint and norm bound; the next section gives numerical tolerances and precision details. Timing includes the two gradient propagations and every line-search trial.

For accuracy references, the stopping criterion uses the convex first-order bound
\begin{equation}
 \frac{J_c(u;y)-J_c(u_\star;y)}{2\|y\|_2^2}
 \leq \frac{g(u)^\top u+\|g(u)\|_1}{\|y\|_2^2}.\label{supp:eq:gap}
\end{equation}
The bound follows by minimizing the linear approximation over the box. We use 64 cases chosen before prediction, with eight per material split, target class and material family. The resulting controls define the accuracy and fixed-control refinement references. Small-grid bounded least-squares, finite-difference gradient and weighted-adjoint checks verify the discrete algebra separately.

On one heterogeneous $64\times64$ benchmark case with 160 time steps, we check the JAX propagator using Gaussian test vectors $u$ and $r$. Automatic differentiation of $r^\top A_cu$ with respect to $u$ agrees with the weighted-adjoint result to a relative $L^2$ difference of $5.44\times10^{-6}$, normalized by the automatic-differentiation result. For the same medium and input $u$, the float32 JAX forward response differs from the independent float64 NumPy recurrence by $3.33\times10^{-6}$ in relative $L^2$ norm.

The A40 benchmark measures the first case in each of the same eight strata in one session. It includes the frozen compact and selected seed-zero models and both seed-zero width-64 paired models. Neural inference uses 20 warmups and 50 synchronized repeats; solvers use three warmups and ten repeats; matrix assembly uses one warmup and five repeats. Tables report the median across cases of within-case medians, their across-case interquartile range (IQR), and mean errors on those same eight cases. Compilation and host transfers are excluded. Float32 paths use highest-precision matrix/convolution settings; errors are independently evaluated by the float64 recurrence. Symmetrized inference batches the two target signs and gives the same predicted control as the two-call implementation. The scan unrolls 16 steps while preserving the recurrence. Matrix-free iteration, cached-matrix iteration and new-material assembly are timed separately; an explicit matrix requires 64\,MiB per material.

\section{Reference Accuracy and Line Search}\label{supp:sec:reference_precision}
Numerical accuracy references use float64 projected FISTA with the conservative step size defined in Section S1.5. We require the normalized half-objective gap in Equation~\eqref{supp:eq:gap} to be below $10^{-5}$, using the NumPy recurrence to evaluate propagation and the gradient. All 64 cases satisfy this criterion at each of the three regularization values in Table~\ref{supp:tab:lambda}. The references at $\lambda=2\times10^{-4}$ also supply the fixed controls for grid refinement. Field errors in these comparisons are evaluated in float64 for both neural and numerical controls.

Timed backtracking FISTA starts at $L=1$, doubles $L$ until quadratic majorization holds, and retains the accepted bound for the next iteration. Its float32 test allows relative tolerance $2\times10^{-6}$ plus absolute tolerance $10^{-12}\|y\|_2^2$. All gradient and trial propagations are included in query timing. Timed solvers start from zero; they do not reuse the certified reference controls.

\section{Additional Refinement and Execution Results}
The main-paper propagation snapshots show the first test case in each of the unseen geometric/layered and reachable/inclusion groups; this index-based selection is independent of prediction quality.

\paragraph{Fine-grid reoptimization.}
We reoptimized ten cases at $n=129$ and two of these at $n=259$, using bilinearly extended targets and prolonged controls. All 12 runs reached 32,768 updates, with float64 gaps of $1.9\times10^{-5}$--$1.6\times10^{-4}$. Mean common-node error on the eight stratified $n=129$ cases was 4.68\%. Fine-grid reoptimization can therefore recover accuracy lost by interpolation, although these finite-budget solves did not reach the $10^{-5}$ stopping threshold.

\paragraph{Batched-sign inference.}
The selected model's canonical two-call inference takes 4.344\,ms, compared with 3.253\,ms when the signs are batched; both implementations produce the same predicted control. A separate terminal-forward evaluation takes approximately 0.479\,ms.

\section{Independent Fixed-Budget Confirmation}\label{supp:sec:budget}
This benchmark compares the trained inverse map with numerical solvers under a short online computation budget. It uses a dedicated target and material distribution, defined below, to evaluate how target structure and speed contrast affect the accuracy available within that budget.

\subsection{Development and Frozen Confirmation}
An 80-case development sweep crosses contrasts 1, 1.5, 2.1 and 3.0 with broad geometry, fine geometry, and reachable controls using 2, 8 or 16 modes per spatial direction (four descriptors per condition). Smooth geometry and low- to mid-frequency reachable targets at moderate contrasts motivate the confirmation study. Fine geometry, high-frequency targets and contrast 3.0 often favor numerical iteration; the complete development results accompany the source. We fixed the confirmation distributions, existing checkpoints, primary budget, baseline set and statistical decision rule before generating the new cases.

Confirmation crosses contrasts $r\in\{1,1.5,2.1\}$ with three target families and 16 paired descriptor IDs. Even IDs use layered media, odd IDs use three inclusions. With $X,Y$ the nodal coordinates, the layered mask is $X>a+b(Y-0.5)$, with $a\sim\mathcal U(0.3,0.65)$ and $b\sim\mathcal U(-0.25,0.25)$. Inclusion centers lie in $[0.2,0.8]^2$ and semiaxes in $[0.08,0.2]$; the mask is their union. Speeds are $\sqrt r$ inside the mask and $1/\sqrt r$ outside. All confirmation speeds lie within the training speed range, although the binary-mask distribution differs from training.

Geometric targets sum two disks of radius 0.12, centers in $[0.25,0.75]^2$ and amplitudes in $[0.5,0.9]$. A Gaussian filter with standard deviation 1.4 grid pixels smooths the sum, followed by the taper $[\sin(\pi X)\sin(\pi Y)]^{1/4}$. Dividing by the larger of one and the peak absolute value bounds the target amplitude. Reachable controls use the first $K\times K$ sine modes, $K\in\{2,8\}$, with independent Gaussian coefficients of standard deviation $1/\sqrt{k^2+l^2}$. Each control is scaled to peak amplitude 0.7 and propagated in NumPy float64 to generate its target. Witness controls are never inverse-training labels.

Material and target random seeds are respectively $2026112901+i$ and $2026113901+i$ for descriptor ID $i=0,\ldots,15$. Reusing each ID across contrasts and target families gives 144 paired cases, rather than 144 independent descriptors. The primary set contains the 96 heterogeneous cases; 48 homogeneous controls are secondary. The benchmark target arrays are distinct from those used in training, validation, the earlier accuracy study, and the development sweep.

The benchmark evaluates the three width-64 models from the capacity confirmation, trained with seeds 0, 1 and 2. Each uses 4,832,705 parameters, 65,536 training pairs, and 60,000 updates. They are evaluated separately; no ensemble prediction or checkpoint reselection is performed. JAX 0.4.30 and Flax 0.8.5 execute the independent A40 benchmark.

\subsection{Solvers and Timing}
Projected gradient, FISTA and gradient-restarted FISTA all solve the same box-constrained regularized problem from zero. The half-objective gradient uses the weighted adjoint in Eq.~\eqref{supp:eq:adjoint}. Backtracking starts at $L=1$, doubles until quadratic majorization holds, and retains the accepted $L$. With extrapolated iterate $v$ and trial displacement $d$, acceptance requires
\[
 \|A_cd\|_2^2+\lambda\|d\|_2^2
 \leq L\|d\|_2^2(1+2\times10^{-6})+10^{-12}\|y\|_2^2.
\]
The restart variant resets momentum when $(v-u_{k+1})^\top(u_{k+1}-u_k)>0$. Every trial is clipped to $[-1,1]$. Backtracking succeeds for every evaluated query.

Measured iteration counts are 3, 4, 6, 8, 12, 16, 24, 32, 64 and 256. For each method and target/contrast condition, timing-only selection chooses the largest measured count whose median latency does not exceed the budget. This rule is fixed before examining confirmation errors. The primary budget is 5\,ms; 10 and 20\,ms are secondary.

Every single query receives one warmup and five synchronized repeats with resident inputs in one A40 session. The median of those repeats defines its latency; the condition median determines budget eligibility. Compilation, transfers, offline training and final-forward verification are excluded. Numerical timings include all propagation, adjoint and backtracking work. The direct network uses its canonical two-call shared-sign projection; the eight-case comparison uses batched signs and the warmup and repeat counts specified in Section S1.5. Their latency values therefore describe different protocols.

Field errors are checked with the independent float64 recurrence. The 4,096-step restarted-FISTA visual references have maximum normalized half-objective gap bound $9.8079\times10^{-5}$. This certificate concerns the regularized objective and differs from the $10^{-5}$ threshold used for the 64 accuracy references in Section S2.

\subsection{Statistics and Secondary Budgets}
For each case, neural results average errors from three separate models, then average cases with equal condition weights. Confidence intervals resample the 16 descriptor IDs with replacement, drawing eight IDs within each material family and preserving all target/contrast pairings. We use 20,000 paired bootstrap replicates with seed 2026114001. Relative reduction is $100(1-\bar e_{\rm neural}/\bar e_{\rm numerical})$ and is recomputed inside each resample. The three training seeds are not counted as additional independent cases.

The 5\,ms advantage persists when any one descriptor cluster is omitted. We remove its six heterogeneous cases together, retain the three-model average and fixed solver iteration counts, and recompute the pooled errors with equal weights for the two material families. Across the 16 deletions, the relative error reduction against FISTA ranges from 43.64\% to 46.23\%, compared with 45.02\% on the full set. This descriptive sensitivity analysis indicates that the pooled improvement is not driven by a single descriptor cluster.

The primary success criterion requires a strictly positive lower 95\% bound against each baseline, positive pooled improvement for every network seed, and condition-level median neural latency within 5\,ms for all seeds. Individual-condition intervals are descriptive and are not adjusted for multiple comparisons.

\begin{figure}[ht]
\centering\includegraphics[width=.9\textwidth]{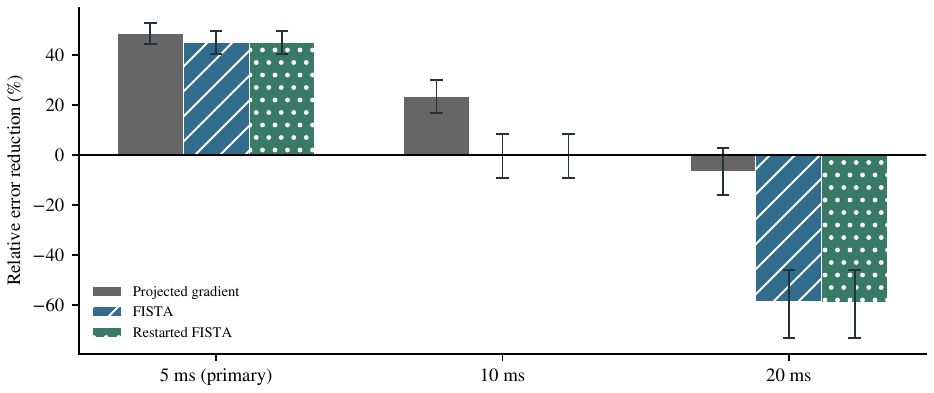}
\caption{Relative field-error reduction by JAX-WAVE over each numerical baseline on the 96 heterogeneous cases. Positive values favor the network. Bars show paired 95\% bootstrap intervals using descriptor-ID resampling stratified by material family. The 5\,ms comparison is primary; 10 and 20\,ms are secondary.}
\label{supp:fig:budget_advantage}
\end{figure}

\begin{table}[ht]
\centering
\caption{Mean field error (\%) on the same 96 heterogeneous confirmation cases. JAX-WAVE predicts once without correction; extra budget is used only by the numerical methods.}
\begin{tabular}{lrrr}
\toprule
Method & 5\,ms & 10\,ms & 20\,ms\\
\midrule
JAX-WAVE & 18.79 & 18.79 & 18.79\\
Projected gradient & 36.51 & 24.48 & 17.64\\
FISTA & 34.18 & 18.71 & 11.84\\
Restarted FISTA & 34.18 & 18.72 & 11.84\\
\bottomrule
\end{tabular}
\end{table}
At 10\,ms, the paired interval for relative error reduction against FISTA is $[-9.24\%,8.22\%]$. At 20\,ms it is $[-73.17\%,-45.79\%]$, favoring FISTA. These results locate the response-budget regime in which direct prediction helps. Figure~\ref{supp:fig:budget_advantage} reports the paired intervals for every baseline and budget.

\begin{figure}[ht]
\centering\includegraphics[width=.9\textwidth]{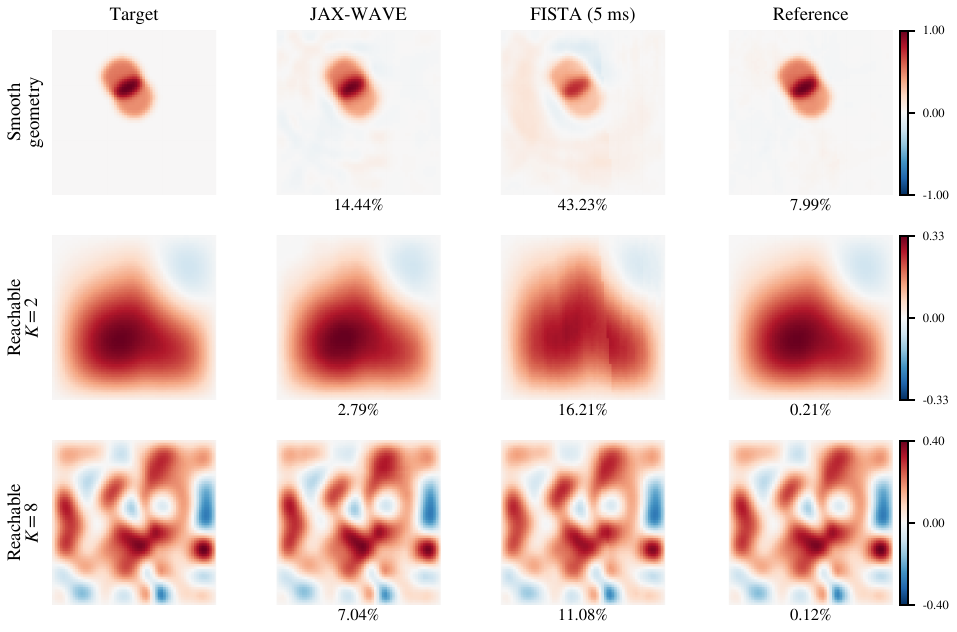}
\caption{Terminal fields for descriptor zero at contrast 1.5 in each confirmation target family (all layered media). Columns show the target, JAX-WAVE seed zero, FISTA within 5\,ms, and the 4,096-step reference. Relative $L^2$ errors are percentages; each row shares a pressure scale. Selection uses descriptor ID, while aggregate comparisons retain all heterogeneous cases and three networks.}
\label{supp:fig:budget_fields}
\end{figure}

\section{Accuracy and Cost for Repeated Queries}
\begin{figure*}[!t]
\centering\includegraphics[width=\textwidth]{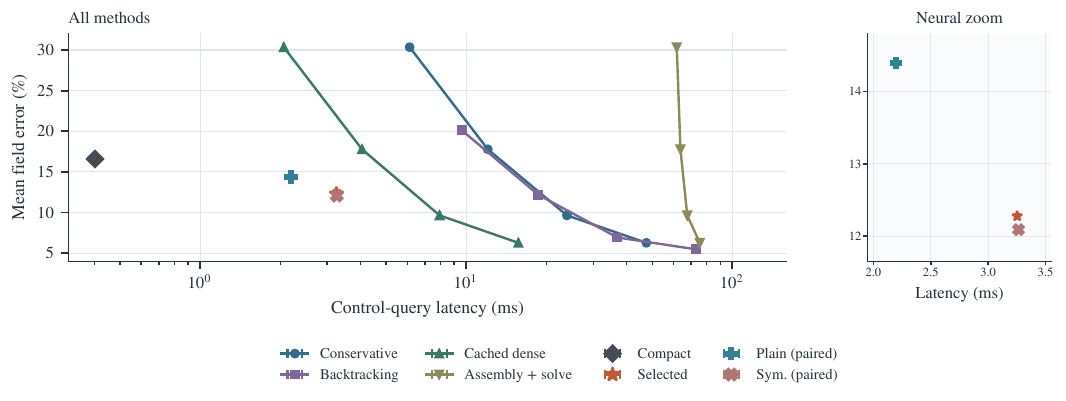}
\caption{Accuracy--latency tradeoffs on eight fixed cases in one A40 session. Points and bars show median and interquartile range of per-case latency medians. The right panel enlarges the three overlapping neural results. Each solver curve uses 8/16/32/64 zero-start FISTA updates from left to right. New-material dense costs add each case's assembly and solve times before aggregation; cached costs exclude assembly. Compilation, transfers, training and final-forward verification are excluded.}
\label{supp:fig:legacy_cost}
\end{figure*}

With sufficient iterations, the polished numerical references achieve lower field error on their matched cases: the certified references achieve unseen-material errors of 14.15\% for geometry and 0.21\% for reachable targets. On exactly the same 16 cases per class, the frozen selected network gives 25.37\% and 4.30\%, averaged over three seeds; the compact model gives 30.58\% and 7.17\%. Known-material numerical errors are 13.52\% and 0.28\%.

\begin{table*}[!t]
\centering
\caption{Same-session A40 control-query comparison on eight stratified cases. Error is the mean over those cases; latency is the median of their synchronized repeat medians, with across-case IQR. Neural rows use seed zero and batched signs where applicable. Cached methods exclude assembly.}
\label{supp:tab:legacy_a40}
\setlength{\tabcolsep}{4pt}
\begin{tabular}{lrrr}
\toprule
Method & Error (\%) & Time (ms) & IQR (ms)\\
\midrule
Frozen compact & 16.59 & 0.402 & 0.402--0.403 \\
Frozen selected & 12.28 & 3.253 & 3.242--3.267 \\
Paired plain & 14.39 & 2.199 & 2.180--2.201 \\
Paired symmetrized & 12.09 & 3.265 & 3.258--3.272 \\
Conservative FISTA, 16 & 17.79 & 12.049 & 12.008--12.068 \\
Conservative FISTA, 32 & 9.65 & 23.903 & 23.858--23.910 \\
Backtracking FISTA, 16 & 12.18 & 18.662 & 18.497--18.890 \\
Backtracking FISTA, 32 & 6.93 & 36.967 & 36.579--37.016 \\
Cached FISTA, 16 & 17.78 & 4.057 & 4.050--4.067 \\
Cached FISTA, 32 & 9.65 & 7.944 & 7.933--7.949 \\
\bottomrule
\end{tabular}
\end{table*}

The latency comparison reveals the benefit of prediction (Table~\ref{supp:tab:legacy_a40}, Figure~\ref{supp:fig:legacy_cost}). On the same eight cases and GPU, the selected network takes 3.253\,ms at 12.28\% error, while 16-step backtracking FISTA takes 18.662\,ms at 12.18\%. A longer numerical solve buys more accuracy: 32-step cached FISTA reaches 9.65\% at 7.944\,ms.

The workload determines whether caching is worthwhile. A new material adds a median 59.773\,ms for matrix assembly (IQR 59.601--59.784\,ms), whereas repeated targets on the same material reuse that matrix. Neural prediction provides short response times in either case. An optional terminal-forward check adds approximately 0.479\,ms.

For a fixed query mix with $t_{\rm num}>t_{\rm net}$, the training cost is amortized after approximately $N_\star=C_{\rm train}/(t_{\rm num}-t_{\rm net})$ queries, with all costs expressed in the same time units. Pairing the A40 symmetrized model's 2,173.1\,s mean optimizer time with its 3.265\,ms query time gives $N_\star\approx1.41\times10^5$ against 16-step backtracking FISTA at 18.662\,ms. Their eight-case errors are close, 12.09\% and 12.18\%. This estimate covers one training run and resident-input queries; data preparation, compilation and model selection add to the cost. Fast response is available after training, while total compute savings depend on query volume and caching.

\section{Compact Symmetry Designs}
At compact scale, we compare the shared sign projection with wider networks, sign augmentation and spatial material conditioning. The projection achieves the lowest validation error and improves both target classes on the original test splits. In the seed-zero comparison, it reduces unseen reachable error from 11.21\% to 8.34\% and geometric error from 34.09\% to 30.64\%. The improvement appears in both material families and on known materials. Table~\ref{supp:tab:accuracy} reports the reconstruction errors for all six models.

\begin{table*}[!t]
\centering
\caption{Mean relative field error (\%) across three training seeds, with sample standard deviation of the per-seed mean. Each run uses the same data and 60,000 updates.}
\label{supp:tab:seeds}
\setlength{\tabcolsep}{4pt}
\begin{tabular}{lrrrr}
\toprule
& \multicolumn{2}{c}{Known materials} & \multicolumn{2}{c}{Unseen materials}\\
Method & Geometry & Reachable & Geometry & Reachable\\
\midrule
U-Net & $33.72\pm0.147$ & $10.95\pm0.070$ & $33.92\pm0.151$ & $11.20\pm0.050$\\
Symmetrized U-Net & $30.17\pm0.144$ & $8.20\pm0.063$ & $30.55\pm0.075$ & $8.26\pm0.070$\\
Spatial odd & $31.30\pm0.004$ & $8.22\pm0.057$ & $31.80\pm0.036$ & $8.24\pm0.045$\\
\bottomrule
\end{tabular}
\end{table*}

The width-20 ordinary U-Net improves accuracy but remains less accurate than the symmetrized width-16 model. Sign augmentation also falls short. Spatial odd improves unseen reachable error over the width-20 control by 2.07 percentage points (material-cluster 95\% interval [1.87,2.28]). The three-seed comparison in Table~\ref{supp:tab:seeds} confirms the symmetrized model's improvement over plain U-Net in every run. Spatial odd and symmetrization have similar reachable errors, with ordering changing across seeds; spatial odd has consistently higher geometric error. These controls support the shared sign projection as a practical choice: it improves geometric reconstruction over spatial modulation while retaining comparable reachable-target accuracy.

\section{Reconstruction-Error Distributions}
\begin{figure*}[!t]
\centering\includegraphics[width=\textwidth]{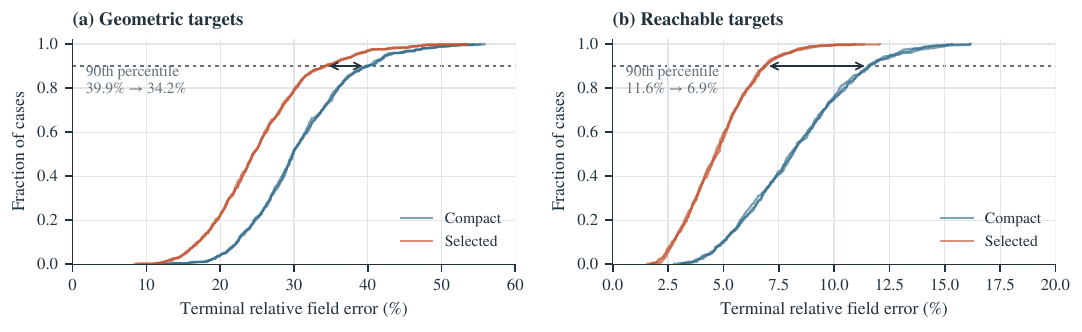}
\caption{Unseen-material confirmation error distributions. Each line is one frozen training seed evaluated on the same 512 cases per target class; colors distinguish compact and selected models. Dashed lines mark the 90th percentile; annotations average the three per-seed percentiles.}
\label{supp:fig:distribution}
\end{figure*}

Figure~\ref{supp:fig:distribution} compares the reconstruction-error distributions of the compact and selected models on unseen materials. The selected model shifts both distributions toward lower errors and reduces the 90th-percentile error in each target class. The improvement thus extends beyond the mean to the high-error portion of the test set.

\section{Supplementary Tables}

\begin{table*}[!htbp]
\centering
\caption{Mean relative field error (\%) on held-out targets. Each material split contains 512 geometric and 512 reachable targets. This six-model comparison uses seed zero, the same training data and 60,000 updates. Bold entries identify the lowest error in each column.}
\label{supp:tab:accuracy}
\begin{tabular}{lrrrrr}
\toprule
& Parameters & \multicolumn{2}{c}{Known materials} & \multicolumn{2}{c}{Unseen materials}\\
Method && Geometry & Reachable & Geometry & Reachable\\
\midrule
U-Net &302,705&33.89&11.02&34.09&11.21\\
Wide U-Net &472,701&32.46&10.06&32.74&10.27\\
Sign augmentation &302,705&34.77&11.13&34.97&11.40\\
Spatial affine &508,176&32.23&9.52&32.56&9.70\\
Spatial odd &477,904&31.30&\textbf{8.16}&31.77&\textbf{8.20}\\
Symmetrized U-Net &302,705&\textbf{30.33}&8.28&\textbf{30.64}&8.34\\
\bottomrule
\end{tabular}
\end{table*}

\begin{table*}[!htbp]
\centering
\caption{Symmetrized-model screening on 256 validation cases, seed zero. $\overline{e^2}$ selects checkpoints; the final columns report mean relative field errors (\%). Independent confirmation is reported in the main paper.}
\label{supp:tab:scaling}
\begin{tabular}{rrrrrrr}
\toprule
$w$ & Training pairs & Updates & Parameters & $\overline{e^2}$ & Geometry & Reachable\\
\midrule
16 & 16,384 & 60,000 & 302,705 & 0.04897 & 29.55 & 7.90 \\
16 & 65,536 & 60,000 & 302,705 & 0.04858 & 29.47 & 7.60 \\
24 & 65,536 & 60,000 & 680,425 & 0.04206 & 27.52 & 6.54 \\
32 & 16,384 & 60,000 & 1,209,057 & 0.04070 & 27.05 & 6.39 \\
32 & 65,536 & 60,000 & 1,209,057 & 0.03833 & 26.33 & 5.87 \\
48 & 65,536 & 60,000 & 2,719,057 & 0.03421 & 24.95 & 5.05 \\
64 & 65,536 & 60,000 & 4,832,705 & 0.03222 & 24.23 & 4.61 \\
16 & 16,384 & 120,000 & 302,705 & 0.04755 & 29.13 & 7.66 \\
32 & 65,536 & 120,000 & 1,209,057 & 0.03696 & 25.90 & 5.52 \\
48 & 65,536 & 120,000 & 2,719,057 & 0.03306 & 24.55 & 4.75 \\
\bottomrule
\end{tabular}
\end{table*}

\begin{table*}[!htbp]
\centering
\caption{Regularization sensitivity on the same 64 reference cases, 32 per target class. Neural results use the symmetrized network trained at each regularization with seed zero; numerical references use the matching objective. Errors are percentages; RMS denotes nodal control root mean square.}
\label{supp:tab:lambda}
\setlength{\tabcolsep}{4pt}
\begin{tabular}{llrrrr}
\toprule
Target & $\lambda$ & \multicolumn{2}{c}{Error (\%)} & \multicolumn{2}{c}{Control RMS}\\
&&Neural&Numerical&Neural&Numerical\\
\midrule
Geometry & $2\times10^{-5}$ & 29.40 & 13.57 & 0.3085 & 0.4488\\
Geometry & $2\times10^{-4}$ & 29.42 & 13.83 & 0.3070 & 0.4300\\
Geometry & $2\times10^{-3}$ & 29.62 & 15.64 & 0.2934 & 0.3757\\
Reachable & $2\times10^{-5}$ & 7.92 & 0.04 & 0.1949 & 0.2030\\
Reachable & $2\times10^{-4}$ & 7.92 & 0.25 & 0.1941 & 0.2008\\
Reachable & $2\times10^{-3}$ & 7.99 & 1.71 & 0.1881 & 0.1932\\
\bottomrule
\end{tabular}
\end{table*}

\begin{table*}[!htbp]
\centering
\caption{Mean field error (\%) in the exploratory material-family probe. Each family contains 64 geometric and 64 reachable targets. All three models retain their original seed-zero weights.}
\label{supp:tab:ood}
\setlength{\tabcolsep}{4pt}
\begin{tabular}{lrrrr}
\toprule
& \multicolumn{2}{c}{Smooth media} & \multicolumn{2}{c}{Six inclusions}\\
Method&Geometry&Reachable&Geometry&Reachable\\
\midrule
U-Net & 41.57 & 16.67 & 35.23 & 12.50\\
Symmetrized U-Net & 39.03 & 13.50 & 31.86 & 9.75\\
Spatial odd & 39.62 & 12.73 & 33.23 & 9.53\\
\bottomrule
\end{tabular}
\end{table*}

\begin{table}[!htbp]
\centering
\caption{Mean relative error (\%) under refined propagation of fixed controls on unseen materials. Each target class contains 16 cases. Errors are evaluated at shared coarse-grid nodes. Numerical controls are obtained for the coarse-grid regularized objective.}
\label{supp:tab:refinement}
\setlength{\tabcolsep}{4pt}
\begin{tabular}{llrrr}
\toprule
Control & Target &64&129&259\\
\midrule
Symmetrized & Geometry &30.72&30.95&31.00\\
Spatial odd & Geometry &31.93&32.15&32.20\\
Numerical & Geometry &14.15&29.18&31.16\\
\midrule
Symmetrized & Reachable &7.20&7.16&7.16\\
Spatial odd & Reachable &7.15&7.15&7.16\\
Numerical & Reachable &0.21&0.94&1.05\\
\bottomrule
\end{tabular}
\end{table}

\begin{table*}[!htbp]
\centering
\caption{Compact architecture-study models and numerical solvers on eight stratified cases in the same A40 session. Time is the median of eight within-case medians, each from five synchronized repeats; the IQR describes across-case variability. Cached methods exclude matrix construction.}
\label{supp:tab:timing}
\setlength{\tabcolsep}{4pt}
\begin{tabular}{lrrr}
\toprule
Method&Error (\%)&Time (ms)&IQR (ms)\\
\midrule
U-Net & 20.56 & 0.409 & 0.407--0.413\\
Symmetrized U-Net & 16.57 & 0.420 & 0.417--0.424\\
Conservative FISTA, 8 & 30.36 & 6.076 & 6.060--6.089\\
Conservative FISTA, 16 & 17.79 & 11.991 & 11.956--12.015\\
Conservative FISTA, 32 & 9.65 & 23.802 & 23.770--23.915\\
Backtracking FISTA, 8 & 20.14 & 9.659 & 9.493--9.848\\
Backtracking FISTA, 16 & 12.18 & 18.771 & 18.583--18.824\\
Backtracking FISTA, 32 & 6.93 & 36.940 & 36.848--37.053\\
Cached FISTA, 16 & 17.78 & 3.997 & 3.983--4.000\\
Cached FISTA, 32 & 9.65 & 7.874 & 7.868--7.880\\
\bottomrule
\end{tabular}
\end{table*}

\end{document}